\documentclass[a4paper, 10pt, conference]{ieeeconf}      
\usepackage{FG2026}
\usepackage{graphicx}
\usepackage{multirow}
\usepackage{url}
\usepackage{hyperref}
\FGfinalcopy 

\IEEEoverridecommandlockouts                              
\def\FGPaperID{72} 

\title{\LARGE \bf
On Applicability of Synthetic Datasets for \\ Facial Expression Recognition
}

\author{\parbox{16cm}{\centering
    {\large Ali Azmoudeh$^1$, Erdi Sarıtaş$^1$, Ömer Yıldırım$^2$, and Hazım Kemal Ekenel$^{1,3}$}\\
    {\normalsize
    $^1$ Department of Computer Engineering, Istanbul Technical University, Istanbul, Türkiye\\
    $^2$ Department of Informatics, University of Zurich, Zurich, Switzerland\\
    $^3$ Division of Engineering, NYU Abu Dhabi, Abu Dhabi, UAE}}
    \thanks{This work was supported by the ITU-Turkcell research scholarship and the Meetween Project that received funding from the European Union's Horizon Europe Research and Innovation Programme under Grant Agreement No. 101135798.
}
}

\usepackage{fancyhdr}
\begin{document}

\ifFGfinal
\thispagestyle{empty}
\pagestyle{empty}
\else
\author{\parbox{16cm}{\centering
    {\large Ali Azmoudeh$^1$, Erdi Sarıtaş$^1$, Ömer Yıldırım$^2$, and Hazım Kemal Ekenel$^{1,3}$}\\
    {\normalsize
    $^1$ Department of Computer Engineering, Istanbul Technical University, Istanbul, Türkiye\\
    $^2$ Department of Informatics, University of Zurich, Zurich, Switzerland\\
    $^3$ Division of Engineering, NYU Abu Dhabi, Abu Dhabi, UAE}}
    \thanks{This work was supported by the ITU-Turkcell research scholarship and the Meetween Project that received funding from the European Union's Horizon Europe Research and Innovation Programme under Grant Agreement No. 101135798.
}
}
\pagestyle{plain}
\fi
\maketitle
\thispagestyle{fancy}

\begin{abstract}

Facial Expression Recognition faces two core challenges. The first is class imbalance in public datasets, which skews the learning process and weakens generalization. The second is related to privacy and data collection constraints, which limit the sharing of facial images and restrict the creation of large, balanced datasets. To address these issues, we examine three complementary strategies for constructing privacy-preserving FER datasets in the standard seven discrete facial expression classes setting. Our strategies are: (i) pseudo-labeling large unlabeled face collections with a teacher model under a confidence-thresholding scheme, (ii) prompt-driven synthesis using diffusion models conditioned on demographic attributes, and (iii) task-aware GAN-based expression editing that modifies facial expression while preserving identity and realism. For training and evaluation, we employed widely adopted datasets, including AffectNet, RAF-DB, and FER2013. We utilized the synthetic datasets DigiFace, DCFace, and EmoNet-Face BIG as unlabeled sources for pseudo-labeling. Additionally, we utilized the FFHQ dataset as the source for generative synthesis. The main experiments are conducted using a classic CNN backbone, IR50, and we also explore a more complex architecture, POSTERv1, to assess its feasibility and robustness. Using cross-dataset evaluations, we analyze the trade-offs each strategy presents in curated datasets. The findings demonstrate how synthetic data can effectively substitute or be combined with real datasets to mitigate imbalance and privacy limitations. \small \textit{Code and generated datasets: \href{https://www.github.com/AliAZ98/SyntFER}{github.com/AliAZ98/SyntFER}}

\end{abstract}

\section{INTRODUCTION}

Facial Expression Recognition (FER) underpins applications in human–computer interaction, affective computing, and behavioral analysis. Despite rapid progress in model capacity, two systemic issues continue to limit real-world implementation. The first is severe class imbalance in public benchmarks~\cite{tutuianu2024benchmarking}, which biases learning toward frequent expressions. The second issue involves privacy~\cite{gdpr,privacyFRsurvey} and data collection~\cite{rafdb,shin2024noisy} constraints~\cite{Liu20253DFD}, which prohibit the sharing of facial imagery and limit the construction of large, demographically balanced datasets~\cite{green2025gender}. These challenges are visible even in widely used in-the-wild datasets, such as AffectNet~\cite{affectnet}, RAF-DB~\cite{rafdb}, and FER2013~\cite{fer2013}. These limitations motivate researchers to reduce dependence on sensitive data while improving diversity control in FER datasets~\cite{fer_survey,taati2025synpain}.

A natural response has been to complement real images with synthetic sources. Synthetic face sets can be rendered at scale with controllable factors, such as pose, illumination, and expression~\cite{digiface}, or synthesized using generative models that disentangle identity from style~\cite{dcface}. At the face recognition level, the community has begun organizing challenges and benchmarks, such as FRCSyn~\cite{synFRchallenge,synFRbenchmark}, and their ongoing evaluation, which explicitly motivate the use of synthetic data to address privacy constraints, annotation costs, and demographic/imbalance issues. A recent synthetic FER dataset, EmoNet-Face~\cite{emonet}, provides fine-grained expression annotations. In parallel, teacher-student pipelines can harvest large unlabeled datasets through pseudo-labeling, in which a strong expert assigns labels with confidence filtering. At the image-level, modern generators enable controllable expression synthesis: text-conditioned latent diffusion, e.g., Stable Diffusion~\cite{stable_diff}, provides semantic guidance for portrait generation, action-unit (AU) aware diffusion offers AU-coherent facial muscle generation, e.g., FineFace~\cite{fineface}, and Generative Adversarial Network (GAN) based editors such as GANmut~\cite{ganmut} modify the expression in a learned, low-dimensional expression space.

\begin{figure}[!t]
    \centering
    \includegraphics[width=0.99\linewidth]{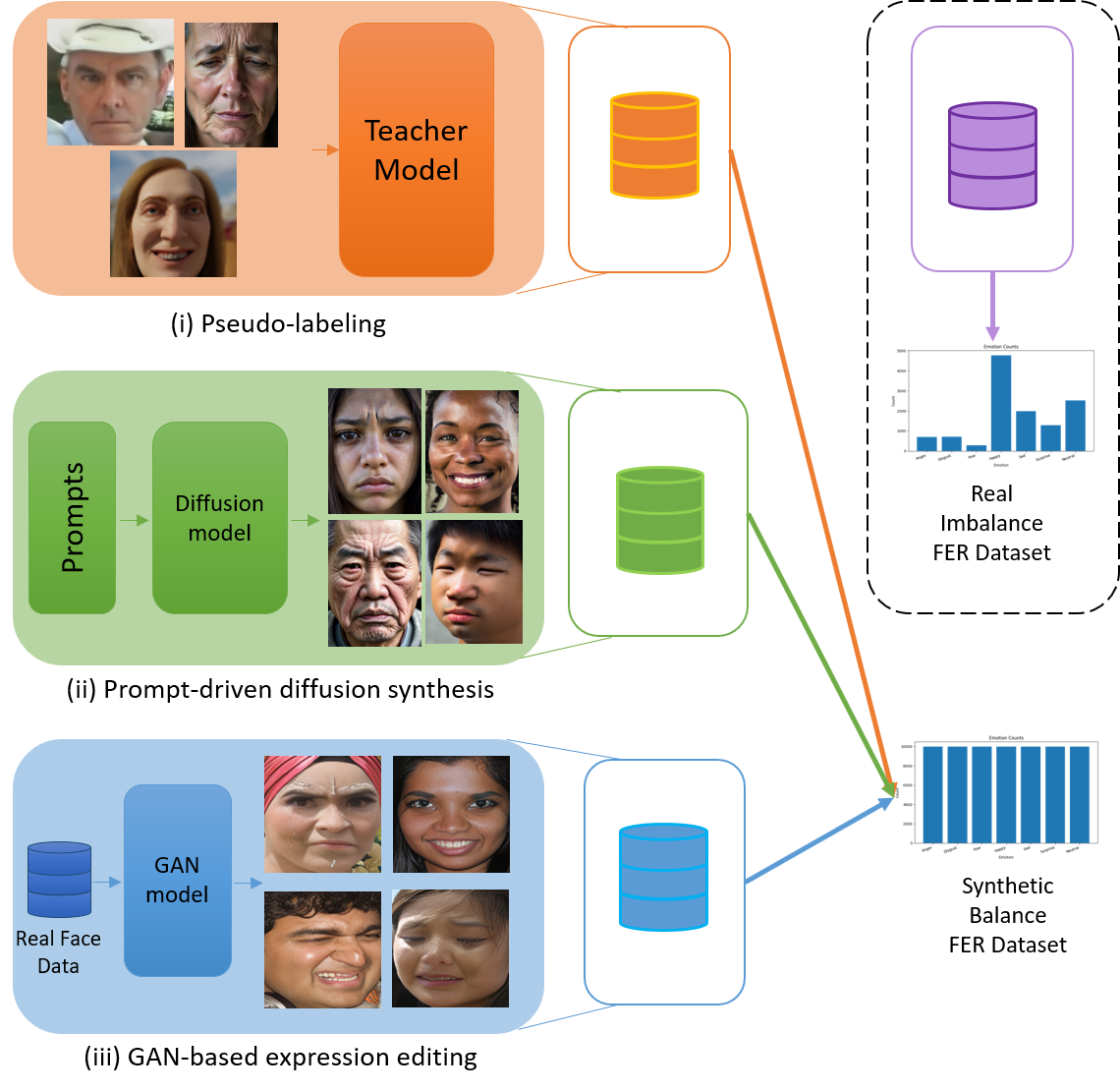}
    \caption{Abstract overview of our FER dataset curation pipelines, with the aim of constructing synthetic balanced datasets. We utilized three synthetic data generation strategies to address class imbalance and privacy constraints.}
    \label{fig:abs-overall}
\end{figure}

This paper examines how these elements can be employed to create privacy-preserving, class-balanced synthetic FER datasets without requiring the collection of sensitive imagery. We study three complementary curation strategies in the seven-class FER setting: (1) \textbf{teacher-guided pseudo-labeling} of large unlabeled sources, DigiFace-1M~\cite{digiface}, DCFace~\cite{dcface}, EmoNet-Face BIG~\cite{emonet}; (2) \textbf{prompt-driven diffusion} for generating new faces with controlled demographics and affect, Stable Diffusion~\cite{stable_diff}, FineFace~\cite{fineface}; and (3) \textbf{task-specific generator} on real faces using GANmut~\cite{ganmut}, with additional restoration. We evaluate these strategies on standard real datasets, AffectNet~\cite{affectnet}, RAF-DB~\cite{rafdb}, and FER2013~\cite{fer2013} using IR50~\cite{arcface} and POSTERv1~\cite{zheng2023poster}.

We focus on the seven basic expressions: anger, disgust, fear, happiness, neutral, sadness, and surprise. Unlabeled sources, such as DigiFace-1M, DCFace, and EmoNet-Face BIG, are used solely for pseudo-labeling or as generation priors, like FFHQ~\cite{ffhq}. Our contributions can be summarized into two folds:
\begin{enumerate}
  \item We present three synthetic FER dataset curation pipelines, including teacher-guided pseudo-labeling, text-conditioned diffusion synthesis, and GAN-based expression editing.
  \item We conduct a systematic comparison by evaluating FER benchmark datasets and analyzing the trade-offs among the three strategies. This includes ablation studies that compare against classical data augmentation and mix the synthetic sources.
\end{enumerate}
\noindent Overview of our study is presented in Fig.~\ref{fig:abs-overall}.

The remainder of this paper is structured as follows. Section~\ref{related} reviews advances in FER learning, dataset limitations, facial image synthesis, and synthetic datasets. Section~\ref{metod} details our methodology, including datasets, classifiers, pseudo-labeling, diffusion-based generation, and GAN-based editing with restoration. Section~\ref{experiment} describes the dataset preparation with training and testing setups. Section~\ref{result} reports the evaluation results and presents our analysis on them. Section~\ref{conclude} concludes our work and discusses limitations and future directions.

\section{RELATED WORKS} \label{related}

We organized the related work to follow the flow of our study. We begin by outlining advances in FER learning and techniques for expression classification. We then examine the challenges of commonly used real FER datasets, such as class imbalance. Next, we survey advances in facial image synthesis that enable controllable, identity-preserving generation and editing, and finally review advances in synthetic datasets that leverage these generative tools to build scalable, privacy-aware resources for FER.

\subsection{Advances in FER Learning}

Methodological progress has paralleled the growth of data~\cite{fer_survey}. Balanced Feature Fusion Network~\cite{li2024bffn} diagnoses class bias in AffectNet/RAF‑DB and introduces synthetic balancing along with an AU‑guided fusion branch to improve fairness without hurting overall accuracy. Leave No Stone Unturned~\cite{zhang2023leave} proposes re‑balanced attention and label smoothing to learn class distributions explicitly. Uncertainty‑Aware Label Distribution Learning~\cite{le2023uncertainty} combines valence–arousal neighbors with a learnable uncertainty factor to model ambiguous labels. Semi‑supervised FER has matured rapidly: ReMoP~\cite{ijcai2025p154} mixes classifier outputs and feature‑similarity distances to generate more reliable pseudo‑labels under skew, regularizing representations into separated categories. Semantic data augmentation~\cite{li2024semantic} perturbs class‑specific latent encodings via a VAE‑GAN~\cite{razghandi2022variational} to enrich rare expressions. RMFER~\cite{cho2024rmfer} constructs reaction‑mashup videos and trains a contrastive learner across neighboring frames to harness temporal cues for underrepresented classes. The 6th ABAW challenge~\cite{abaw} solution by Yu et al.~\cite{yu2024exploring} couples pseudo‑label‑based pretraining with a temporal encoder to boost robustness on in‑the‑wild data. These works collectively confirm that label reweighting, feature‑space augmentation, temporal contrastive learning, and balanced sampling are complementary tools for mitigating class imbalance. In addition, noisy/ambiguous labels are another obstacle in FER and have motivated noise-tolerant training objectives~\cite{gu2022toward,shin2024noisy}. Architectural developments for FER models have also been studied, including mixing global and local features for long‑tailed recognition~\cite{zhou2023mixing}, a pyramid cross‑fusion transformer network~\cite{zheng2023poster}, and the cross‑attention transformer network~\cite{mao2025poster++}. Cross-dataset generalization has also been explored: several studies~\cite{bhati2025generalized,huang2023fer,li2024emotion,ji2021region} train on RAF-DB and evaluate on the AffectNet validation set, reporting accuracies in the 41.57\%--58.43\% range, with the best result achieved by NorFace~\cite{liu2024norface}.

\subsection{Challenges of Real FER Datasets}

Early FER datasets, such as CK+~\cite{ckplus} and FEW~\cite{few}, contained only a few thousand images and were collected under controlled environments. Even large-scale in-the-wild datasets, such as AffectNet~\cite{affectnet}, RAF-DB~\cite{rafdb}, and FER-2013~\cite{fer2013}, suffer from severe class imbalance, noisy annotations, and limited demographic diversity~\cite{tutuianu2024benchmarking,fer_survey,rafdb,shin2024noisy}. Recent analyses have highlighted that collecting privacy-sensitive data~\cite{gdpr} and obtaining reliable labels are challenging, which limits the size and quality of datasets and hinders the training of deep-learning models~\cite{privacyFRsurvey,shin2024noisy}.

\subsection{Advances in Facial Image Synthesis}

Bozorgtabar et al.~\cite{expression_analysis} propose an attribute-guided face synthesis and domain-adaptation framework for FER that conditions the generator on face-landmark heatmaps and leverages landmark-based frontalization. GANmut~\cite{ganmut} learns a continuous, interpretable expression space in which direction corresponds to expression and radius to intensity, enabling smooth morphing among basic expressions. While GAN-based methods have long served as a strong framework for identity-preserving synthesis, diffusion models are increasingly favored over them~\cite{dcface}. FineFace~\cite{fineface} injects Facial Action Unit vectors into a frozen diffusion backbone via an adapter path, enabling localized and intensity‑controlled expression generation. Arc2Face~\cite{arc2face} proposes a foundation model that generates faces directly from ArcFace~\cite{arcface} embeddings, thereby avoiding the need for textual prompts and providing a compact identity handle. ID‑Booth~\cite{idbooth} employs a triplet identity objective to ensure identity consistency during diffusion‑based generation. In addition, DiffusionAct~\cite{diffusionact} introduces a one-shot diffusion autoencoder that transfers head pose and expressions without subject-specific fine-tuning. IP‑FaceDiff~\cite{ipfacediff} adapts text‑to‑image diffusion models to facial video editing, using targeted fine‑tuning and identity preservation losses to achieve high‑quality, localized edits at a fraction of the computational cost. 

\subsection{Advances in Synthetic Datasets}

Graphics-rendered and generative datasets offer a means to sidestep consent and demographic limitations. DigiFace-1M~\cite{digiface} utilizes 3D digital humans to generate millions of faces with programmable control over pose, illumination, expression, and accessories. Furthermore, SynFace~\cite{synface} combines real and synthetic identities using GANs, whereas DCFace~\cite{dcface} employs dual-condition diffusion to separate identity from style, thereby achieving both intra-class diversity and identity fidelity. At the face recognition level, community benchmarks such as FRCSyn~\cite{synFRchallenge,synFRbenchmark} and their ongoing evaluations explicitly motivate the use of synthetic data to address privacy, labeling, and demographic imbalance concerns. EmoNet-Face BIG~\cite{emonet} pushes further by releasing over 200k synthetic faces annotated with 40 fine-grained expressions and balanced demographics. VariFace~\cite{variface} demonstrates that fairness can be enhanced with diffusion-based synthetic datasets using a two-stage pipeline that incorporates demographic label refinement and diversity guidance. SynFER~\cite{synfer}, a framework for synthesizing expression images, couples textual prompts with facial action units, semantic guidance, and a pseudo-label generator to produce diverse, realistic expressions paired with reliable labels.

\section{METHODOLOGY} \label{metod}

\begin{figure*}[!t]
  \centering
  \includegraphics[width=0.8\textwidth]{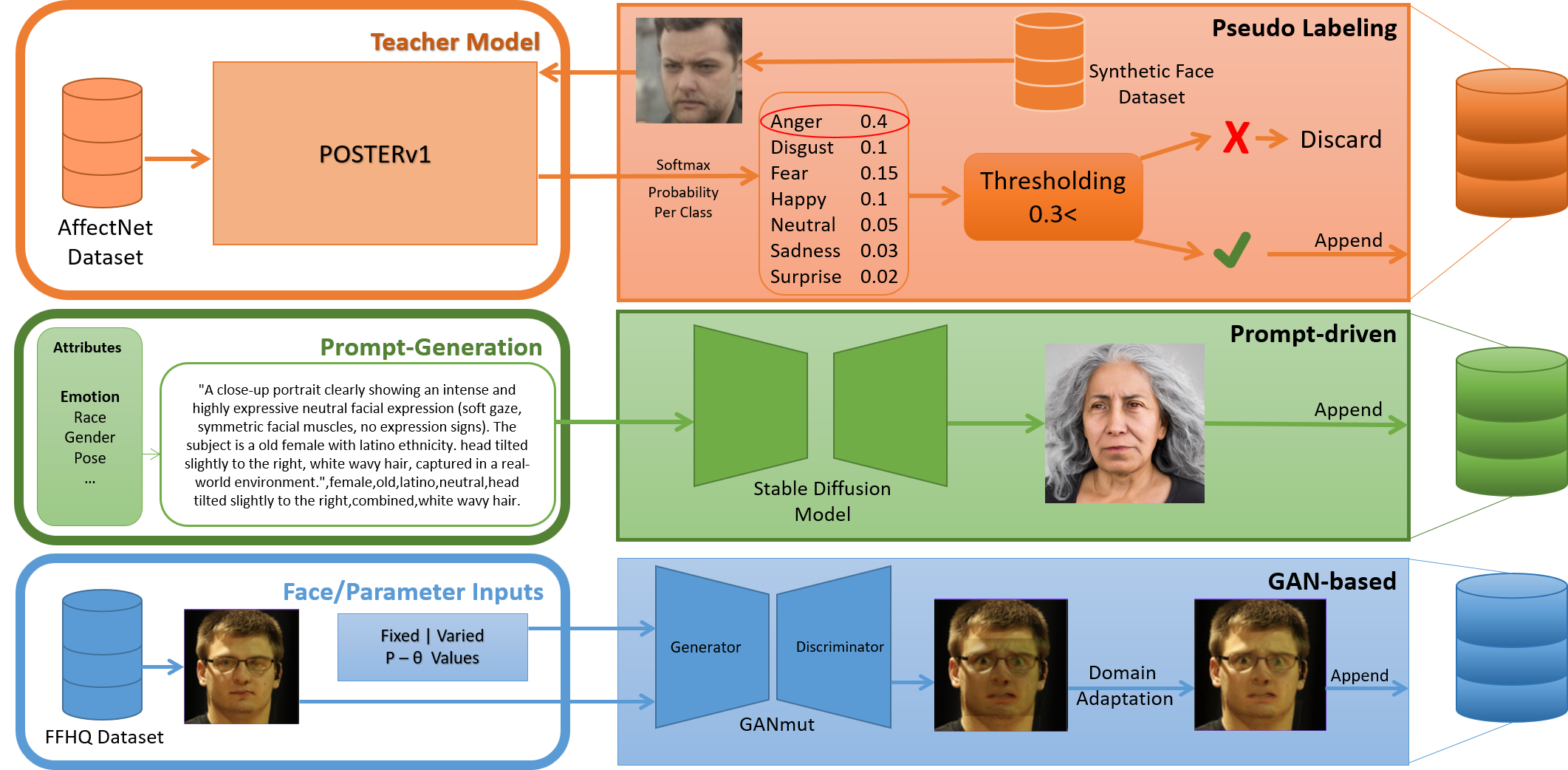}
\caption{Overview of our dataset curation pipelines. \textbf{Top (Pseudo-labeling):} a POSTERv1 teacher infers class posteriors on unlabeled synthetic faces; samples with max softmax probability $\geq 0.3$ are kept and others discarded, and the highest-confidence samples are selected. \textbf{Middle (Prompt-driven):} attribute tuples (expression, race, gender, etc.) drive diffusion models to generate photorealistic portraits. \textbf{Bottom (GAN-based):} real faces are edited with GANmut using polar codes $(\rho,\theta)$ and then adapted to the source (real-image) domain via color transfer, alpha blending, and restoration with CodeFormer.}
  \label{fig:overall_data}
\end{figure*}

In this section, we first describe the datasets and baseline FER classifiers used in our study. We then examine three synthetic dataset curation pipelines: (i) pseudo-labeling of large unlabeled pools, (ii) diffusion-based, prompt-driven face synthesis, and (iii) task-specific GAN editing for expression control. The general overview of each pipeline is presented in Fig.~\ref{fig:overall_data}. With these techniques, we aim to address class imbalance while minimizing reliance on sensitive real images. We follow the seven basic expressions: anger, disgust, fear, happiness, neutral, sadness, and surprise.

\subsection{Datasets}

We used AffectNet~\cite{affectnet}, RAF-DB~\cite{rafdb}, and FER2013~\cite{fer2013} as real FER datasets. DigiFace-1M~\cite{digiface}, DCFace~\cite{dcface}, and EmoNet-Face BIG~\cite{emonet} serve as unlabeled sources for pseudo-labeling. FFHQ~\cite{ffhq} provides the source images for expression modification.

\begin{itemize}
    \item \textbf{AffectNet}~\cite{affectnet} includes a large collection of in-the-wild face images annotated with both categorical labels and valence–arousal values. It contains more than 1M $224\times224$ RGB face images, of which approximately 450k are manually labeled. In our study, we used approximately 281K samples from the seven basic expression categories. This dataset was used to train the teacher model for pseudo-labeling. Additionally, we used the validation set containing 3{,}500 samples, equally distributed across all expression classes, for cross-dataset evaluation.

    \item \textbf{RAF-DB}~\cite{rafdb} provides in-the-wild face images labeled with the seven basic expressions. We followed the standard split: approximately 12K samples for training and 3{,}000 for testing. The images are in $100\times100$ RGB format. Both the training and test sets were utilized in our experiments.

    \item \textbf{FER2013}~\cite{fer2013} consists of low-resolution grayscale face images captured in the wild. The image resolution is $48\times48$. We followed the default dataset split, which includes approximately 28K training samples, 3{,}500 public test samples, and 3{,}500 private test samples. For cross-dataset evaluation, we combined the public and private test sets to form a test set of about 7{,}000 images.

    \item \textbf{DigiFace-1M}~\cite{digiface} is a synthetic dataset made up of 3D-rendered faces with programmatic control over pose, expression, lighting, accessories, and camera parameters. The images are aligned using facial landmarks. The dataset contains approximately 1.2M RGB images at a resolution of $112\times112$.

    \item \textbf{DCFace}~\cite{dcface} is a diffusion-based synthetic dataset that employs dual conditioning mechanisms—an identity prior and a real-image style bank—to balance identity consistency and intra-class diversity. The dataset is available in different scales, ranging from approximately 0.5M to 1.2M images, all in $112\times112$ RGB format. We used the 0.5M version in our work.

    \item \textbf{EmoNet-Face}~\cite{emonet} is a synthetic text-to-image dataset for affective expression generation. It consists of three subsets: BIG (approximately 200K images with LLM-assisted labels), BINARY (about 20K images with more than 65K binary human annotations), and HQ (2{,}500 images with 10K expert-rated intensity scores). The dataset is demographically balanced. In our study, we use the BIG subset, which includes $512\times512$ RGB images. When we refer to EmoNet-Face in the manuscript, we specifically mean this BIG version.

    \item \textbf{FFHQ}~\cite{ffhq} is a dataset of real face images at $1024\times1024$ resolution, consisting of 70K RGB samples. It offers a wide range of ages, ethnicities, and accessories. All images are carefully filtered and aligned.
\end{itemize}

\subsection{FER Classifiers}

\paragraph{IR50~\cite{arcface}}
We used IR50 (iResNet-50), a face-focused ResNet-50 variant pretrained on the MS-Celeb-1M~\cite{msceleb} dataset for face recognition. It takes $112\times112$ RGB inputs and is pretrained with the ArcFace loss. The final classification layer is removed; the model outputs a 1024-d embedding. This improves stability and discriminability in face recognition settings. To adapt the FER model to 7 expression classifications, we replaced the model's head with a linear classifier with 7 classes.

\paragraph{POSTERv1~\cite{zheng2023poster}}
To test our data on a higher-capacity model, we also employed a Convolutional Neural Network (CNN)-Transformer hybrid architecture, POSTERv1. It runs two streams in parallel, IR50 image features and landmark features (from a frozen MobileFaceNet~\cite{mobilefacenet}), and fuses them through a cross-fusion transformer with query swapping. A three-level feature pyramid (512/256/128) with depth-8 transformer encoders improves robustness to scale changes.

\subsection{Pseudo Labeling}

We apply pseudo-labeling to DigiFace-1M~\cite{digiface}, DCFace~\cite{dcface}, and EmoNet-Face~\cite{emonet}. A POSTERv1~\cite{zheng2023poster} model trained on AffectNet~\cite{affectnet} acts as the teacher/expert, where the trained model achieved $66.41\%$ accuracy on the validation set of the AffectNet dataset~\cite{azmoudeh2024advanced}. For each image, we compute softmax class probabilities and assign the class with the highest probability. To exclude uncertain samples, we retain only those with a probability of the assigned class greater than or equal to a threshold; otherwise, we discard them. We selected a threshold of 0.3 to balance sample quantity and model confidence (more than twice the probability of random selection over seven classes). We manually reviewed the assigned samples based on different probability scores and found an equilibrium that accounted for the sample sizes in each class. We also retain the probability values to select the highest-confidence samples. The retained high-confidence predictions serve as pseudo-labels for downstream training. 

\subsection{Face Generation with Diffusion Models}

We generate faces from text prompts using two diffusion models: (1) Stable Diffusion~\cite{stable_diff} (Realistic-Vision v5.1) for general photorealistic portraits, and (2) FineFace~\cite{fineface} for AU-aware, expression-controlled synthesis.

\paragraph{General Stable Diffusion}

We instantiate a Stable/Latent Diffusion pipeline~\cite{stable_diff} that denoises in the VAE latent space and conditions each step on the text-prompt embedding via cross-attention. We access a photorealistic portrait checkpoint (Realistic-Vision v5.1~\footnote{\url{https://huggingface.co/stablediffusionapi/realistic-vision-v51}})
through the HuggingFace Diffusers API~\cite{diffusersAPI}.

To build a balanced and reproducible prompt set, we enumerate a controlled factor space and convert each combination (the Cartesian product) into a concise, readable sentence that specifies demographics and the target expression. Each prompt starts with a fixed portrait style as “A close-up, in-the-wild portrait \ldots”, then names the target expression, specifies demographics -- age, gender, and race/ethnicity -- adds a concise expression cue, and sets a mild head pose plus a single neutral identity trait like “glasses” to increase diversity without coupling to the expression. Scenes use natural lighting and a plain background to stabilize alignment and training. Expression cues and identity traits are chosen consistently with the target expression and the corresponding age--gender group.

\begin{itemize}
\item \textbf{Gender} $\in$ \{male, female\}
\item \textbf{Age} $\in$ \{child, young, adult, middle-aged, older\}
\item \textbf{Race/ethnicity} $\in$ \{White, Black, Asian, Middle-Eastern, Latino\}
\item \textbf{Head pose} $\in$ \{frontal, slight left/right yaw, slight up/down pitch\}
\end{itemize}

Expression cues use two formats: descriptive phrases such as “subtle smile”, “raised brows”, FACS-style cues -- textual description of units like AU6{+}AU12~\cite{zhi2020comprehensive} are used -- and also a combined version of both -- similar to FACS-style but with a more informal structure. Expression cues and identity traits are constructed using GPT-5~\cite{gpt5}.

We use the following template: “A close-up portrait clearly showing an intense and highly expressive $<$expression$>$ facial expression ($<$expression\_cues$>$). The subject is a $<$age$>$ $<$gender$>$ with $<$race$>$ ethnicity. $<$head\_pose$>$, $<$identity\_trait$>$, captured in a real-world environment.”. For the expression cue, we use all three versions. For the identity trait, we choose neutral details such as “wearing glasses” or “short beard.” A complete example: “A close-up portrait clearly showing an intense and highly expressive disgusted facial expression (upper lip lifted with nose wrinkling and narrowed eyes). The subject is a female child of Latino ethnicity. Slightly turned to the right, short bob cut, captured in a real-world environment.”.  All the prompts with their used flags are stored in a CSV file. Prompts in each row from this CSV file are fed to the Stable Diffusion model.

\begin{table*}
\centering
\caption{Per-expression sample counts for the training sets.}
\label{tab:per_expression_counts_base}
\scriptsize
\begin{tabular}{|c||ccccccc|}
\hline
\textbf{Dataset} & \textbf{Angry} & \textbf{Disgust} & \textbf{Fear} & \textbf{Happy} & \textbf{Sad} & \textbf{Surprise} & \textbf{Neutral} \\ \hline
RAF-DB & 705 & 717 & 281 & 4{,}772 & 1{,}982 & 1{,}290 & 2{,}524 \\
\hline
DCFace & 10{,}000 & 9{,}572 & 2{,}039 & 10{,}000 & 10{,}000 & 10{,}000 & 10{,}000 \\ 
DigiFace & 10{,}000 & 10{,}000 & 10{,}000 & 10{,}000 & 10{,}000 & 10{,}000 & 10{,}000 \\
EmoNet-Face & 10{,}000 & 2{,}201 & 10{,}000 & 10{,}000 & 10{,}000 & 10{,}000 & 10{,}000 \\
\hline
Stable Diffusion & 6{,}307 & 5{,}913 & 6{,}015 & 7{,}280 & 6{,}249 & 7{,}008 & 6{,}716 \\
FineFace & 4{,}627 & 4{,}013 & 4{,}839 & 4{,}342 & 5{,}038 & 5{,}189 & 4{,}499 \\
FineFacev2 & 5{,}437 & 4{,}754 & 5{,}926 & 4{,}939 & 5{,}729 & 6{,}181 & 5{,}315 \\
\hline
GANmut-F & 9{,}958 & 9{,}935 & 9{,}952 & 9{,}951 & 9{,}953 & 9{,}967 & 9{,}950 \\
GANmut-V & 9{,}958 & 9{,}937 & 9{,}954 & 9{,}952 & 9{,}958 & 9{,}966 & 9{,}950 \\
\hline
\end{tabular}
\end{table*}

\paragraph{AU-aware FineFace}

We employ an AU-controlled diffusion model, FineFace~\cite{fineface}, which injects continuous Facial Action Unit (AU) intensity vectors into a text-conditioned latent diffusion backbone. Text preserves identity and scene context, while AUs steer facial musculature and control intensity, thereby disentangling expression from prompt wording and yielding reproducible “recipes” that pair well with our demographic/pose prompts. The model uses both the text prompt and AUs as conditioning factors in the diffusion process. We tried two versions of prompts for the FineFace.

\begin{itemize}
    \item \textbf{v1:} We create a new subject-only prompt by filling the template with the flags, deliberately excluding any expression words: “A close-up portrait of a \texttt{<age>} \texttt{<gender>} with \texttt{<race>} ethnicity, \texttt{<identity\_trait>}, \texttt{<head\_pose>}, photorealistic, natural lighting.”. The expression is supplied solely via the AU vector obtained from the fixed Expression $\rightarrow$ AU mapping.
    \item \textbf{v2:} We begin from the full prompt produced in the previous (General Stable Diffusion) stage, but first remove any expression-related phrases. We then compute the AU vector from the target expression. We also append a compact, human-readable AU clause to the end of the cleaned prompt. 
\end{itemize}

\subsection{Expression Manipulation with GANs}

\begin{figure}[!t]
    \centering
    \includegraphics[width=0.5\linewidth]{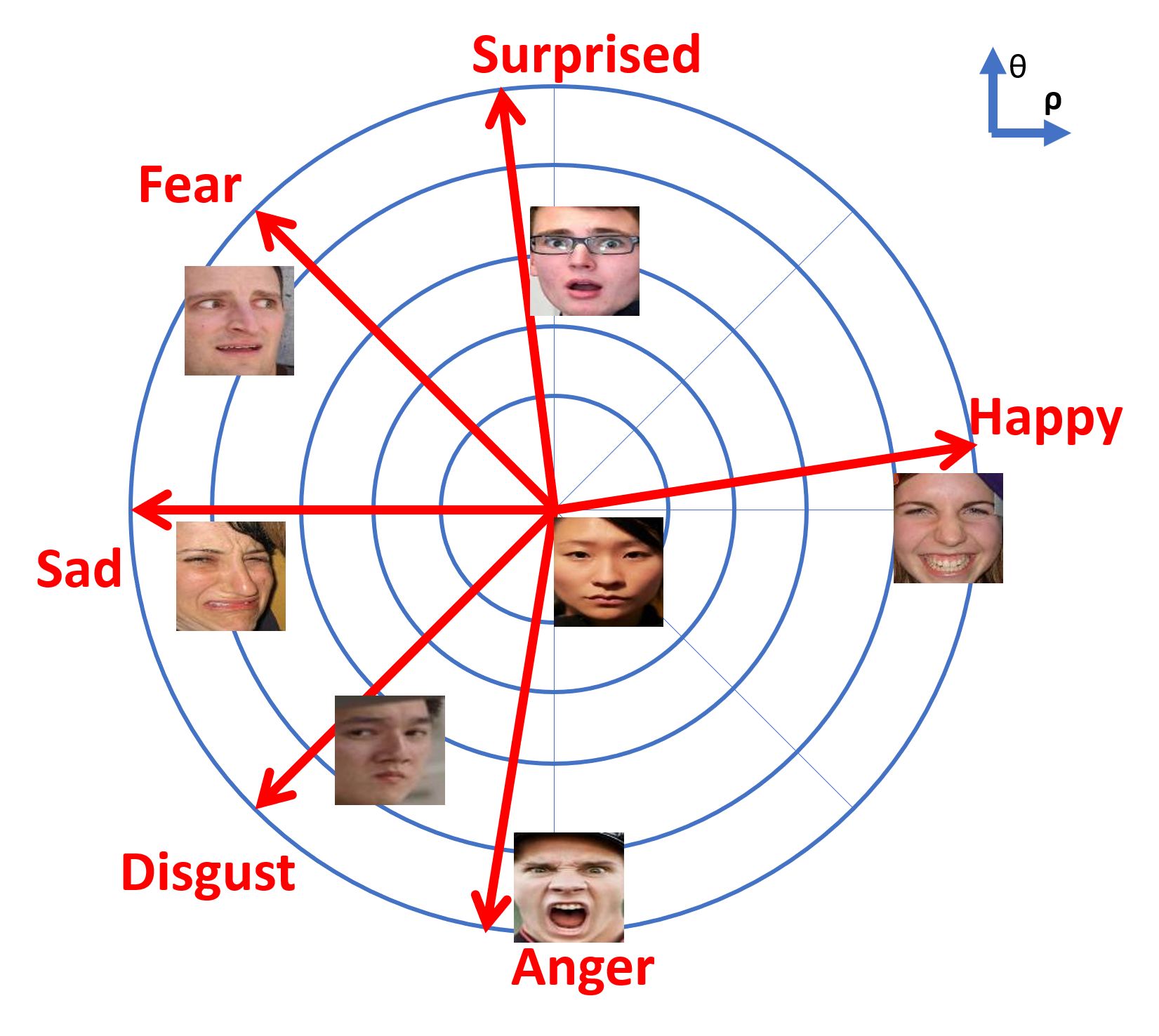}
    \caption{Illustration of the GANmut polar coordinate system.}
    \label{fig:ganmut-ro}
\end{figure}

To edit expressions on real faces, e.g., from FFHQ~\cite{ffhq}, we use GANmut~\cite{ganmut}, a task-specific GAN. Instead of using one-hot labels, GANmut learns a low-dimensional control space in polar form. This polar coordinate is illustrated in Fig.~\ref{fig:ganmut-ro}, where the angle selects the basic expression and the radius controls intensity. Interpolating directions produces compound expressions. Training follows a StarGAN-style multi-domain setup~\cite{stargan} with a Wasserstein objective with gradient penalty~\cite{wgan}, plus (i) a domain classification loss, (ii) an InfoGAN-inspired coordinate-regression term to link images to their 2D code~\cite{infogan}, (iii) a monotonic-intensity regularizer for coherent trajectories, and (iv) a cycle/reconstruction constraint to preserve identity and pose~\cite{cyclegan}.

\paragraph{Sampling policies}
We use two regimes. \emph{Fixed-intensity} sampling selects codes at (or near) each expression’s canonical direction (or Gaussian mode mean), producing prototypical edits with consistent intensity. \emph{Variate-intensity} sampling moves along the radius to vary intensity and interpolates between directions for blends; in the Gaussian case, we add small perturbations around each mode. In practice, we set the coordinates (for both Fixed and Variate versions) based on manual observations to cover standard and nuanced expressions.

\begin{figure}[!t]
    \centering
    \includegraphics[width=0.9\linewidth]{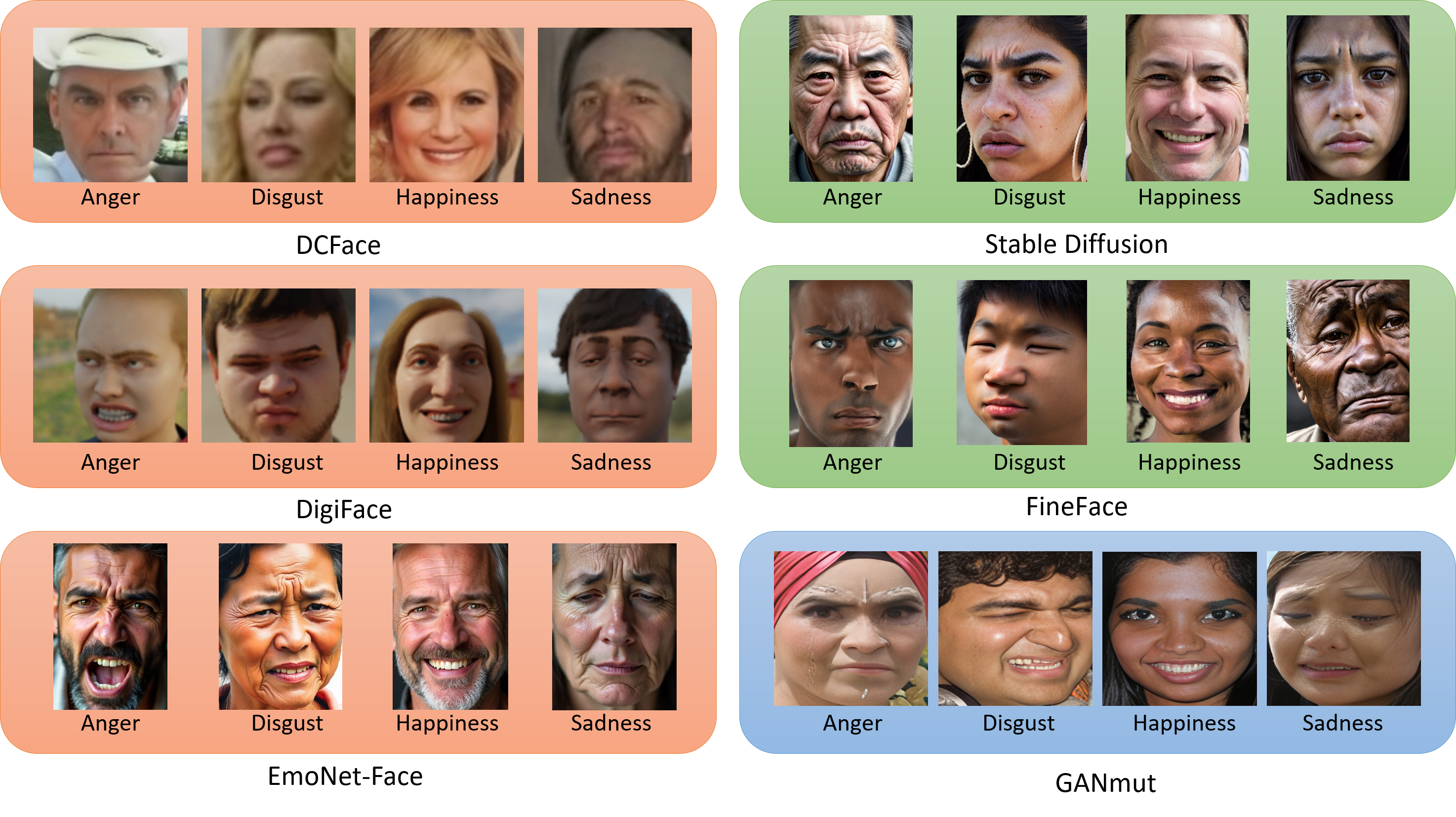}
    \caption{Generated face image samples from the constructed datasets. The expression labels are given below the samples.}
    \label{fig:dataset_samples}
\end{figure}

\paragraph{Compositing and blending}
Edits are produced on aligned crops and then pasted back into the originals. To reduce shifts and boundary artifacts, we first apply color transfer in CIELAB to match means and variances between the edited region and its local context, then convert to RGB. We then use alpha blending with a soft-border mask to suppress stitching artifacts and maintain a photorealistic appearance.

\paragraph{Refinement with CodeFormer}
As the final step, we apply CodeFormer~\cite{codeformer} to enhance the visual quality of the modified images. CodeFormer performs blind face restoration as discrete code prediction from a learned VQ codebook~\cite{vqcodebook}, with a transformer predictor and a fixed decoder. To prevent the restorer from amplifying artifacts introduced by GANmut editing, we first apply degradation, with particular emphasis on the paste boundary. Concretely, for the face box $(x,y,w,h)$ we draw a thin rectangular ring mask of width $t$ (30 pixels) and apply a two-scale blur+noise operator: a stronger \emph{local} Gaussian blur (kernel size of 15 with sigma value of 5) and higher-variance (sigma value of 0.15) noise inside the ring, composed with a lighter \emph{global} (kernel size of 5 with sigma value of 2) Gaussian blur and low-variance (sigma value of 0.05) noise over the full image. This smooths discontinuities at the boundary while preserving interior facial detail, so CodeFormer focuses on holistic restoration rather than inadvertently enhancing artifacts.

\section{EXPERIMENTAL SETUP} \label{experiment}

\subsection{Dataset Preparation}

We organized three training regimes using the synthetic sources, including DCFace, DigiFace, EmoNet-Face, Stable Diffusion–driven data, FineFace, and GANmut-generated data. Full synthetic sets are constructed per source as class-balanced collections, with at most 10K images per expression (a total of at most 70K samples per dataset). The primary reason for this choice is to minimize generation and training time.  Moreover, in addition to these full-synthetic datasets, we created hybrid datasets by combining them with the RAF-DB, following two settings: \emph{Concatenation (C)} and \emph{Fix (F)}. For the Concatenation setting, we take the union of RAF-DB (train) and all samples from the chosen synthetic source of the same class, without enforcing a per-class limit of 10K. In the Fix setting, we target a per-class total of no more than 10{,}000 images. For each expression class \(c\), we start from RAF-DB (train) and add samples from a single synthetic source until the class reaches 10K. For Concatenation, our goal is to minimize data loss, whereas for Fix, we aim to achieve a more balanced hybrid dataset.

Across all regimes, we aimed to achieve equal per-class splits to address class imbalance in FER. Except for the DCFace and EmoNet-Face datasets, the generated datasets are balanced. Moreover, except for the already-aligned DCFace and DigiFace images, all face images are preprocessed with MTCNN~\cite{mtcnn}, ensuring consistent framing and geometry across datasets and a shared evaluation protocol. Samples from each curated dataset are presented in Fig.~\ref{fig:dataset_samples} (more examples can be found in the Supplementary Material). However, some faces could not be used because MTCNN failed to detect them. The number of images per class of the training sets (including RAF-DB) is given in Table~\ref{tab:per_expression_counts_base}. Regarding the constructed datasets, more information on realism, including Fr\'echet Inception Distance (FID)~\cite{heusel2017gans} and Kernel Inception Distance (KID)~\cite{binkowski2018demystifying} scores using CLIP~\cite{radford2021learning} vision embeddings with the CelebA-HQ~\cite{karras2017progressive} dataset as the target, as well as diversity encompassing gender, race, and age predictions inferred by FaceXFormer~\cite{narayan2025facexformer}, can be found in the Supplementary Material.

To rigorously evaluate the contribution of synthetic data properties against simple data quantity, and to assess the complementarity of different generators, we constructed two additional training sets:

\paragraph{\textit{Balancing with Traditional Augmentation}}
    Assessing whether performance improvements arise solely from class balancing instead of synthetic diversity, we create a dataset derived from RAF-DB that is real and class-balanced. We employed a classical augmentation pipeline including random horizontal flips, rotation ($\pm 10^{\circ}$), affine shear ($\pm 10^{\circ}$), translation ($10\%$), scaling ($0.85\times-1.15\times$), and photometric jitter (brightness, contrast, saturation). Using the "original + augmented" strategy, we add augmented samples to the minority classes in RAF-DB to match the majority class's sample count, ensuring balanced distribution. We named this version "Add-on". Additionally, we create a purely augmented dataset with a class-wise sample size of 10{,}000, which we call "All-aug".

\paragraph{\textit{Mixed Synthetic Ensemble Dataset}}
We curated a unified dataset that combines samples from all considered synthetic sources (DCFace, DigiFace, EmoNet-Face, Stable Diffusion, FineFace (V1 and V2), and GANmut (F and V)) to explore the synergy between different generative approaches. To maintain a manageable dataset size and ensure balance, we sampled 1,250 images per expression class from each synthetic source, yielding a balanced dataset of 10{,}000 images per class. We named this dataset "Mixed-SYN". Similar to other synthetic datasets, we also combined it with RAF-DB (C version). Further, we conducted all-synthetic training using a synthetic validation set rather than a real one. The synthetic validation set is collected in a manner similar to the training set, but at a 1/10 ratio. We identify this result as "Mixed-SYN*".

\subsection{Training and Testing}
We train two backbones, IR50 and POSTERv1, under two regimes: full-synthetic and hybrid (real data plus synthetically generated datasets), consistent with the seven-class FER setting. Due to model initialization from pre-trained weights, we adopt a low learning rate varied per dataset in the range \(1 \times 10^{-6}\) to \(1 \times 10^{-7}\) with a batch size equal to 64. Optimization uses Sharpness-Aware Minimization (SAM)~\cite{foret2020sharpness} with the Adam optimizer, and the loss function is a combination of weighted (\(\lambda\)) label-smoothing cross-entropy (with \(\lambda = 2\)) and standard cross-entropy. All experiments are conducted on a single NVIDIA RTX 3080 Ti GPU. We trained the models for 120 epochs, saving the best-performing model from the epoch where the RAF-DB test set served as the validation objective.

Evaluation is conducted on three widely used FER benchmarks -- RAF-DB, AffectNet, and FER2013 -- employing RAF-DB as the sole training source and primary indicator of in-domain validation performance, while FER2013 and AffectNet (test sets) serve as the principal benchmarks for assessing cross-dataset generalization. Performance is reported using overall Accuracy and F1-score, with model selection for each training run based strictly on the best performance on the RAF-DB test set. Notably, cross-dataset test sets -- AffectNet and FER2013 -- are never utilized for hyperparameter tuning or model selection. This setup is designed to rigorously examine whether synthetic data can effectively replace or complement real data to yield more generalizable models, especially since only the AffectNet validation set is balanced.

\section{RESULTS}  \label{result}

\begin{table}[t!]
\centering
\caption{Cross-dataset performance of IR50 (Accuracy / F1-Score).}
\label{tab:ir50_results}
\footnotesize
\setlength{\tabcolsep}{2pt} 
\begin{tabular}{|l|c||c|c|c|}
\hline
\multirow{2}{*}{\textbf{Source}} & \textbf{RAF-DB} & \multicolumn{3}{c|}{\textbf{Acc / F1 (\%)}} \\ \cline{3-5} 
 & \textbf{Usage} & \textbf{RAF-DB} & \textbf{FER2013} & \textbf{AffectNet} \\
\hline \hline
RAF-DB & - & 89.60 / 83.03 & 50.79 / 41.93 & 40.20 / 35.82 \\ 
\hline
\multirow{3}{*}{DCFace} 
 & $\times$ & 32.92 / 22.96 & 36.26 / 33.16 & 38.07 / 36.91 \\
 & C & 89.70 / 83.07 & 51.83 / 46.20 & 50.73 / 48.39 \\
 & F & 89.77 / 83.22 & 51.48 / 46.03 & 50.90 / 48.63 \\ 
\hline
\multirow{3}{*}{DigiFace} 
 & $\times$ & 41.69 / 29.54 & 39.12 / 36.68 & 37.87 / 35.16 \\
 & C & \textbf{90.22} / \textbf{83.85} & 53.76 / 47.45 & 48.06 / 45.59 \\
 & F & 89.90 / 83.66 & 54.28 / 47.92 & 48.26 / 45.59 \\ 
\hline
\multirow{3}{*}{EmoNet} 
 & $\times$ & 56.52 / 46.29 & 41.61 / 37.03 & 52.63 / 50.48 \\
 & C & 88.79 / 82.34 & 53.61 / 46.63 & 53.89 / 52.43 \\
 & F & 89.09 / 82.39 & 53.32 / 46.24 & \textbf{54.87} / \textbf{53.42} \\ 
\hline
\multirow{2}{*}{Stable Diff} 
 & $\times$ & 57.53 / 35.05 & 38.49 / 30.87 & 30.38 / 23.06 \\
 & C & 88.27 / 80.14 & 54.44 / 45.22 & 43.50 / 39.08 \\ 
\hline
\multirow{2}{*}{FineFace} 
 & $\times$ & 50.55 / 33.86 & 38.31 / 30.34 & 33.68 / 29.98 \\
 & C & 88.33 / 80.13 & 54.90 / 45.76 & 43.93 / 39.56 \\ 
\hline
\multirow{2}{*}{FineFace v2} 
 & $\times$ & 47.88 / 34.84 & 35.39 / 30.87 & 30.92 / 28.98 \\
 & C & 88.33 / 80.20 & 54.93 / 45.78 & 43.93 / 39.56 \\ 
\hline
\multirow{3}{*}{GANmut-F} 
 & $\times$ & 62.39 / 48.28 & 47.06 / 40.42 & 51.54 / 50.04 \\
 & C & 89.73 / 82.28 & \textbf{55.92} / 48.04 & 51.74 / 49.62 \\
 & F & 89.57 / 82.37 & 55.82 / \textbf{48.11} & 51.94 / 49.83 \\ 
\hline
\multirow{3}{*}{GANmut-V} 
 & $\times$ & 59.35 / 50.12 & 42.78 / 38.91 & 53.03 / 52.75 \\
 & C & 88.95 / 81.80 & 54.96 / 47.29 & 52.91 / 51.55 \\
 & F & 89.11 / 82.11 & 54.61 / 47.69 & 54.09 / 52.48 \\ 
\hline
\multicolumn{5}{l}{$^{*}$\footnotesize{$\times$ all synthetic, C for Concatenation and F for Fix RAF-DB usage tags.}}
\end{tabular}
\end{table}

This section analyzes the effectiveness of synthetic data in facial expression recognition, comparing performance in in-domain and cross-domain scenarios. Tables~\ref{tab:ir50_results} and~\ref{tab:poster_results} present the performance of IR50 and POSTERv1, respectively. The columns report the \textit{Accuracy / F1-Score} pairs. Additionally, Tables~\ref{tab:ir50_aug} and~\ref{tab:ir50_mix} present ablation studies for classical augmentation and mixed synthetic sources. Lastly, to investigate a detailed performance comparison, we report class-wise accuracies in Table~\ref{tab:class-wiseIR50}.

\subsection{IR50 Results}

Our baseline on the real RAF-DB dataset is shown in Table~\ref{tab:ir50_results}, achieving 89.60\% accuracy and 83.03\% F1-score. We observe that purely synthetic training (marked with $\times$) generally underperforms compared to real data, highlighting the persistent domain gap. However, combining synthetic data with real data (Concatenation 'C' and Fixed 'F' settings) consistently yields the most robust models. As shown in Table~\ref{tab:ir50_results}, hybrid models outperform the real-only baseline on out-of-domain targets. For instance, GANmut-F (C) improves FER2013 accuracy to 55.92\%, and EmoNet-Face (F) leads on AffectNet with 54.87\% accuracy.  Additionally, we observe a trade-off in sampling policies for the GANmut variants: Fixed-Intensity (GANmut-F) sampling favors alignment with datasets like FER2013, while Variable-Intensity (GANmut-V) sampling better captures the broad variability of AffectNet.

\begin{table}[t!]
\centering
\caption{Cross-dataset performance of IR50 model -- Augmentation Ablation.}
\label{tab:ir50_aug}
\footnotesize
\setlength{\tabcolsep}{4pt}
\begin{tabular}{|l||c|c|c|}
\hline
\multirow{2}{*}{\textbf{Usage}} & \multicolumn{3}{c|}{\textbf{Acc / F1 (\%)}} \\ \cline{2-4} 
 & \textbf{RAF-DB} & \textbf{FER2013} & \textbf{AffectNet} \\
\hline \hline
Add-on  & 90.03 / 84.93 & 49.72 / 45.43 & 48.18 / 46.65 \\
All-aug & 90.74 / 85.82 & 58.18 / 49.53 & 51.19 / 48.04 \\
\hline
\end{tabular}
\end{table}

\begin{table}[t!]
\centering
\caption{Cross-dataset performance of IR50 model -- Mixed Synthetic Sources.}
\label{tab:ir50_mix}
\scriptsize
\setlength{\tabcolsep}{4pt}
\begin{tabular}{|l||c|c|c|}
\hline
\multirow{2}{*}{\textbf{Source}} & \multicolumn{3}{c|}{\textbf{Acc / F1 (\%)}} \\ \cline{2-4} 
 & \textbf{RAF-DB} & \textbf{FER2013} & \textbf{AffectNet} \\
\hline \hline
Mixed-SYN* & 60.66 / 50.85 & 48.31 / 44.71 & 56.50 / 56.33 \\
Mixed-SYN   & 60.72 / 50.99& 48.66 / 45.01& 56.53 / 56.38\\
Mixed-SYN-C & 88.62 / 81.20 & 54.85 / 49.64 & 57.02 / 56.36 \\
\hline
\end{tabular}
\end{table}

A key question is whether performance gains stem from synthetic diversity or merely increased data volume. Table~\ref{tab:ir50_aug} addresses this by comparing classical augmentation against our synthetic strategies. The All-aug version improves in-domain and out-of-domain performance compared to the standard RAF-DB training, outperforming all FER2013 results. However, the Add-on strategy shows less improvement, with a slight decrease in FER2013 accuracy.

Table~\ref{tab:ir50_mix} explores the synergy between different generative approaches. The Mixed-SYN experiment demonstrates that combining diverse synthetic sources significantly closes the domain gap, achieving 56.50\% accuracy on AffectNet, surpassing the real RAF-DB baseline. Furthermore, the Mixed-SYN-C ensemble pushes this boundary further, achieving the highest overall generalization scores on AffectNet across all experiments (57.02\% Accuracy / 56.36\% F1). Additionally, Mixed-SYN* shows that even the absence of a real validation set can be approximated with a synthetic one, with only a slight performance loss.

\begin{table*}[h]
\centering
\caption{Class-wise performance of IR50 Model on AffectNet validation set.}
\label{tab:class-wiseIR50}
\footnotesize
\begin{tabular}{|c||rrrrrrr|}
\hline
\multirow{2}{*}{\textbf{Source}} & \multicolumn{7}{c|}{Classes}                                                                                         \\ \cline{2-8} 
 &
  \multicolumn{1}{c}{\textbf{Anger}} &
  \multicolumn{1}{c}{\textbf{Disgust}} &
  \multicolumn{1}{c}{\textbf{Fear}} &
  \multicolumn{1}{c}{\textbf{Happiness}} &
  \multicolumn{1}{c}{\textbf{Neutral}} &
  \multicolumn{1}{c}{\textbf{Sadness}} &
  \multicolumn{1}{l|}{\textbf{Surprise}} \\ \hline
RAF-DB                           & 17.87          & 19.48          & 9.04           & 91.94          & 65.26          & 45.78          & 36.22          \\
All-aug                          & 23.9           & \textbf{47.19} & 16.87          & 91.73          & 77.71          & 51.2           & 52.31          \\
DCFace (F)                       & 24.3           & 44.58          & 23.69          & 90.52          & 54.22          & \textbf{67.07} & 50.5           \\
DigiFace (F)                     & 25.3           & 32.93          & 25.1           & 93.15          & 58.63          & 62.65          & 40.24          \\
EmoNet-Face (F)                  & 29.92          & 49.8           & 41.37          & 89.72          & 76.31          & 35.94          & \textbf{61.97} \\
Stable Diffusion (C)             & 21.49          & 18.27          & 10.64          & \textbf{94.35} & 74.5           & 39.76          & 47.89          \\
FineFace (C)                     & 21.69          & 19.48          & 9.84           & \textbf{94.35} & 74.1           & 40.76          & 47.48          \\
FineFace v2 (C)                  & 21.69          & 19.48          & 9.84           & \textbf{94.35} & 74.1           & 40.76          & 47.48          \\
GANmut-F (F)                     & 36.55          & 31.73          & 20.08          & 92.34          & 79.72          & 48.8           & 54.53          \\
GANmut-V (F)                    & \textbf{38.96} & 30.32          & 34.14          & 90.52          & 78.31          & 55.02          & 51.51          \\
Mixed-SYN-C                      & 36.14          & 45.18          & \textbf{45.18} & 89.31          & \textbf{79.92} & 47.79          & 55.73          \\ \hline
\end{tabular}%
\end{table*}

Performance varies by data source type. Table~\ref{tab:class-wiseIR50} details the class-wise accuracy on the AffectNet validation set. We observe that diffusion-based sources (Stable Diffusion, FineFace) produce high-fidelity images that maintain strong performance on dominant classes (e.g., Happiness $\sim$94\%) but struggle with minority classes (e.g., Fear $\sim$10\%), compared to GAN-based methods ($\sim$20-34\%). Consequently, GANmut variants offer better class balance. Additionally, we observe that mixing data sources does not introduce a performance imbalance; rather, it preserves it.

\subsection{POSTERv1 Results}

\begin{table}[b!]
\centering
\caption{Cross-dataset performance of POSTERv1 (Accuracy / F1-Score).}
\label{tab:poster_results}
\footnotesize
\setlength{\tabcolsep}{2pt}
\begin{tabular}{|l|c||c|c|c|}
\hline
\multirow{2}{*}{\textbf{Source}} & \textbf{RAF-DB} & \multicolumn{3}{c|}{\textbf{Acc / F1 (\%)}} \\ \cline{3-5} 
 & \textbf{Usage} & \textbf{RAF-DB} & \textbf{FER2013} & \textbf{AffectNet} \\
\hline \hline
RAF-DB & - & \textbf{91.33} / \textbf{86.13} & 52.48 / 44.42 & 47.43 / 44.95 \\ 
\hline
\multirow{2}{*}{DCFace} 
 & C & 90.81 / 54.34 & 53.06 / 47.71 & 54.18 / 52.32 \\
 & F & 90.91 / 84.55 & 53.69 / 48.26 & 54.09 / 52.24 \\ 
\hline
\multirow{2}{*}{DigiFace} 
 & C & 90.84 / 84.48 & 56.28 / \textbf{49.64} & 52.48 / 50.34 \\
 & F & 90.97 / 84.41 & 56.12 / 49.56 & 52.57 / 50.41 \\ 
\hline
\multirow{2}{*}{EmoNet} 
 & C & 90.55 / 84.10 & 56.03 / 49.32 & 55.33 / 54.07 \\
 & F & 90.71 / 84.56 & 56.06 / 49.44 & 55.44 / \textbf{54.17}\\ 
\hline
Stable Diff & C & 89.47 / 81.87 & 55.59 / 46.55 & 44.56 / 40.07 \\ 
\hline
FineFace & C & 89.50 / 82.26 & 56.52 / 47.56 & 46.80 / 43.64 \\ 
\hline
FineFace v2 & C & 89.28 / 82.68 & 54.28 / 44.68 & 44.73 / 41.72 \\ 
\hline
\multirow{3}{*}{GANmut-F} 
 & $\times$ & 41.33 / 30.59 & 31.89 / 37.20 & 30.55 / 31.89 \\
 & C & 90.71 / 84.72 & 56.65 / 49.50 & 54.09 / 52.47 \\
 & F & 91.00 / 84.95 & \textbf{56.88} / 49.21 & 53.72 / 51.98 \\ 
\hline
\multirow{3}{*}{GANmut-V} 
 & $\times$ & 41.33 / 30.59 & 31.93 / 27.24 & 30.55 / 31.89 \\
 & C & 89.96 / 83.42 & 55.70 / 48.44 & 54.67 / 53.62 \\
 & F & 90.19 / 83.50 & 55.59 / 48.36 & \textbf{54.75} / 53.77\\ 
\hline
\end{tabular}
\end{table}

To ensure our findings are not architecture-specific, we validated them using the POSTERv1 backbone. Table~\ref{tab:poster_results} confirms that the trends remain consistent: hybrid training consistently improves cross-dataset generalization regardless of the underlying network architecture. Notably, the GANmut-F (F) model achieves 56.88\% accuracy on FER2013, and GANmut-V (F) reaches 54.75\% on AffectNet, significantly outperforming the RAF-DB baseline.

\section{CONCLUSIONS}\label{conclude}

This work investigates whether synthetic data can systematically address two persistent bottlenecks in Facial Expression Recognition: class imbalance and privacy-constrained data collection without relying on additional sensitive imagery. We instantiated three complementary curation strategies: (i) teacher-guided pseudo-labeling of large unlabeled synthetic pools, (ii) demographic-aware, prompt-driven diffusion synthesis, and (iii) task-specific GAN-based expression editing. We employed IR50 as a simpler model for our general analysis, while POSTERv1 served as a more complex model to evaluate the impacts of generated datasets across different model types. We conducted experiments in the seven-class FER setting, using RAF-DB as the in-domain test set and FER2013 and AffectNet as out-of-domain targets. Performance was reported as accuracy and F1 Scores. Training regimes comprised synthetic-only (\texttt{$\times$}) and two hybrid variants with RAF-DB (Concatenation \texttt{C} and Fix \texttt{F}).

From the results, we observe that synthetic data can serve as a valuable source of training data. Particularly, it can improve cross-dataset generalization when combined with real data. Our ablation studies further confirm that these gains stem from the semantic diversity of synthetic data rather than mere data quantity, as synthetic hybrids outperformed classical augmentation baselines. Additionally, the performance loss in the in-domain setting is minimal and may even lead to an improvement. No single synthetic family dominates all targets, suggesting practical value in designing sources to benchmarks. Curation type has a major impact, particularly when it is synthetic-only, as with RAF-DB, which uses results that are relatively close to each other. Moreover, the dataset size and its diversity may also be directly related to the characteristics of the target test set. The small (diffusion-based) and less diverse (GANmut-F) datasets performed better on RAF-DB and FER, while the others achieved better scores on AffectNet.

\paragraph*{\textbf{Discussion \& Future Work}}
We limit the sample size per class to 10K to balance the training time. The diffusion-based datasets are smaller due to the trade-off between the number of prompts and the divergence of their images. Future work can explore different datasets, potentially hybrid versions including AffectNet, and training methods. A limitation of this study is the reliance on a single teacher model for pseudo-labeling, which may introduce class-dependent label bias; future research will consider alternative teachers and calibration-aware sampling.

{\small
\bibliographystyle{ieee}

\bibliography{egbib}

}


\onecolumn
\section*{Supplementary Materials}
\subsection*{Additional Figures} \label{sec: app 1}

\begin{figure}[h]
  \centering
  \includegraphics[width=\linewidth]{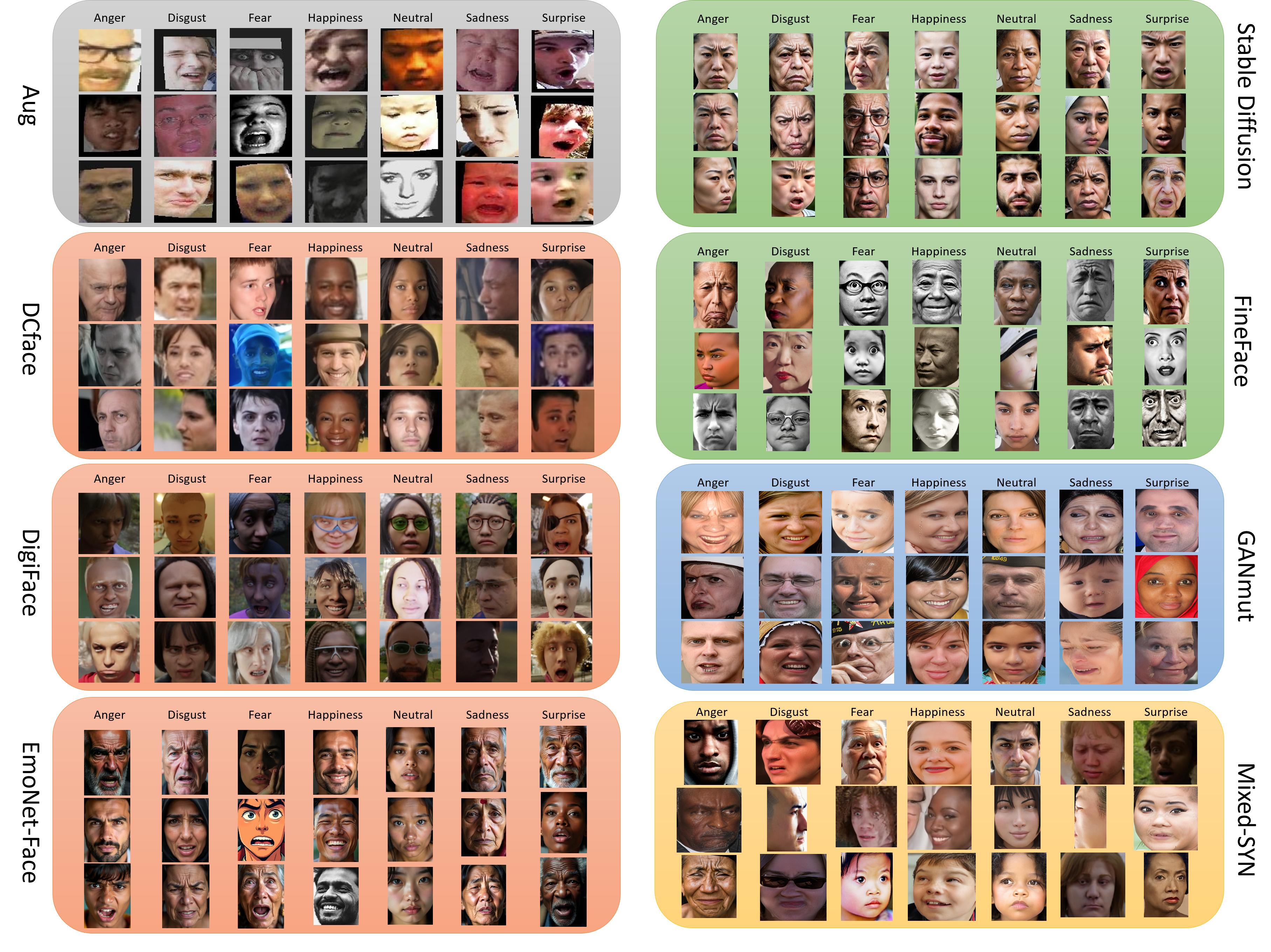}
  \caption{Larger representation of samples in curated datasets}
  \label{fig:app:moremsaple}
\end{figure}

\onecolumn

\subsection*{Additional Tables} \label{sec: app 2}

\begin{table*}[h]
\centering
\caption{Aggregated Fr\'echet Inception Distance (FID) and Kernel Inception Distance (KID) metrics (reference: CelebA-HQ), with FFHQ and RAF-DB included as real-world baselines, and demographic statistics for synthetic datasets.}
\label{tab:agg_grouped_transposed}
\footnotesize
\setlength{\tabcolsep}{3pt}
\begin{tabular}{|c|c|rrrrrrrrrr|}
\hline
\multicolumn{2}{|c|}{\textbf{Category}} & \rotatebox{45}{\textbf{FFHQ}} & \rotatebox{45}{\textbf{RAF-DB}} & \rotatebox{45}{\textbf{DCFace}} & \rotatebox{45}{\textbf{DigiFace}} & \rotatebox{45}{\textbf{EmoNet-Face}} & \rotatebox{45}{\textbf{Stable Diffusion}} & \rotatebox{45}{\textbf{FineFace}} & \rotatebox{45}{\textbf{FineFace v2}} & \rotatebox{45}{\textbf{GANmut-F}} & \rotatebox{45}{\textbf{GANmut-V}} \\ \hline
\multicolumn{2}{|c|}{\textbf{Total}} & 70,000 & 12,271 & 61,611 & 70,000 & 62,201 & 45,488 & 32,547 & 38,281 & 69,666 & 69,675 \\ \hline
\multirow{2}{*}{\textbf{Metrics}} & FID  & 19.29 & 37.67 & 22.18 & 41.30 & 30.72 &  34.54& 27.83 & 29.66 & 25.19 & 24.67 \\
 & KID  & 0.0394 & 0.0926 & 0.0584 & 0.1082 & 0.0576 &  0.0706& 0.0566 & 0.0626 & 0.0570 & 0.0556 \\
\hline
\multirow{2}{*}{\textbf{Gender}} & Male  & - & - & 35,795 & 40,655 & 30,420 & 22,558 & 15,662 & 20,447 & 35,878 & 35,009 \\
 & Female  & - & - & 25,816 & 29,345 & 31,781 & 22,930 & 16,885 & 17,834 & 33,788 & 34,666 \\
\hline
\multirow{5}{*}{\textbf{Race}} & White  & - & - & 45,270 & 44,630 & 22,018 & 9,182 & 14,141 & 14,322 & 55,844 & 56,512 \\
 & Black  & - & - & 3,469 & 10,860 & 7,851 & 6,219 & 3,311 & 4,258 & 2,402 & 2,634 \\
 & Indian  & - & - & 4,213 & 5,349 & 13,843 & 9,088 & 5,429 & 8,956 & 2,972 & 3,523 \\
 & Asian  & - & - & 169 & 997 & 5,942 & 994 & 1,350 & 1,040 & 1,088 & 897 \\
 & Others  & - & - & 8,490 & 8,164 & 12,547 & 20,005 & 8,316 & 9,705 & 7,360 & 6,109 \\
\hline
\multirow{8}{*}{\textbf{Age}} & 0-9  & - & - & 149 & 566 & 2,252 & 4,657 & 5,699 & 4,629 & 7,454 & 7,713 \\
 & 10s  & - & - & 12,074 & 23,602 & 18,283 & 17,674 & 10,141 & 17,234 & 28,986 & 30,653 \\
 & 20s  & - & - & 20,026 & 31,986 & 10,421 & 5,080 & 5,892 & 3,614 & 12,944 & 12,466 \\
 & 30s  & - & - & 8,267 & 6,421 & 4,074 & 2,042 & 2,265 & 2,286 & 10,294 & 9,022 \\
 & 40s  & - & - & 16,858 & 5,637 & 3,560 & 8,065 & 3,732 & 4,276 & 7,108 & 6,864 \\
 & 50s  & - & - & 3,568 & 915 & 9,106 & 5,809 & 3,007 & 4,179 & 2,536 & 2,487 \\
 & 60s  & - & - & 609 & 583 & 6,277 & 1,839 & 1,335 & 1,281 & 272 & 387 \\
 & 70+  & - & - & 60 & 290 & 8,228 & 322 & 476 & 782 & 72 & 83 \\
\hline
\end{tabular}
\end{table*}

\end{document}